\documentclass[9pt]{article}
\usepackage{spconf,amsmath,graphicx}
\usepackage{amssymb}
\usepackage{enumitem}
\usepackage{multirow}
\usepackage{url}

\usepackage[belowskip=-10pt,aboveskip=2pt]{caption}
\ninept
\newlength\myindent
\def\Width{0\kern2\tabcolsep\ldots\kern1\tabcolsep0}
\usepackage{etoolbox}
\newcommand{\zerodisplayskips}{%
  \setlength{\abovedisplayskip}{3pt}
  \setlength{\belowdisplayskip}{3pt}
  \setlength{\abovedisplayshortskip}{3pt}
  \setlength{\belowdisplayshortskip}{3pt}}
\appto{\normalsize}{\zerodisplayskips}
\appto{\small}{\zerodisplayskips}
\appto{\footnotesize}{\zerodisplayskips}
\title{Audio-to-Intent Using Acoustic-Textual Subword Representations \\from End-to-End ASR}
\name{Pranay Dighe, Prateeth Nayak, Oggi Rudovic, Erik Marchi, Xiaochuan Niu, Ahmed Tewfik}
\address{Apple}

\begin{document}
\maketitle
\vspace{-2mm}
\begin{abstract}
\vspace{-1mm}
Accurate prediction of the user intent to interact with a voice assistant (VA) on a device (e.g. a smartphone) is critical for achieving naturalistic, engaging, and privacy-centric interactions with the VA. To this end, we present a novel approach to predict the user intention (whether the user is speaking to the device or not) directly from acoustic and textual information encoded at subword tokens which are obtained via an end-to-end (E2E) ASR model. Modeling directly the subword tokens, compared to modeling of the phonemes and/or full words, has at least two advantages: (i) it provides a unique vocabulary representation, where each token has a semantic meaning, in contrast to the phoneme-level representations, (ii) each subword token has a reusable ``sub''-word acoustic pattern (that can be used to construct multiple full words), resulting in a largely {\it reduced} vocabulary space than of the full words. To learn the subword representations for the audio-to-intent classification, we extract: (i) \textit{acoustic information} from an E2E-ASR model, which provides frame-level CTC posterior probabilities for the subword tokens, and (ii) \textit{textual information} from a pretrained continuous bag-of-words model capturing the semantic meaning of the subword tokens. The key to our approach is that it combines acoustic subword-level posteriors with text information using the notion of positional-encoding to account for {\it multiple} ASR hypotheses simultaneously. 
We show that the proposed approach learns robust representations for audio-to-intent classification and correctly mitigates $93.3\%$ of unintended user audio from invoking the VA at $99\%$ true positive rate.
\end{abstract}

\vspace{-2mm}
\section{Introduction}
\vspace{-2mm}
\label{sec:intro}
In typical voice-assistant (VA) architectures on devices like smartphones, any input audio is first gated by a wake-word detection module, which actively listens for a wake-word (e.g. ``Hey Siri'', ``Hey Alexa'', ``Okay Google'', and so on). It only allows audio anchored with the wake-word to be processed by the downstream models. This gating mechanism is often referred to as the user intent classification (but other names such as  false-trigger-mitigation as well as device-directed-speech-detection are interchangeably used).

Prior work is mainly focused on key-word spotting and wake-up word detection. These approaches typically rely on multi-stage neural network based processing of acoustic features to determine the presence of the wake-word~\cite{sigtia2018vt,kumatani2017direct,wu2018monophone,guo2018time,norouzian2019exploring}. Despite the fact that the latest wake-word detectors are relatively highly accurate, they can still confuse unintended speech as intended for device. Such false alarms have adverse effects on the user engagement and overall experience as well as privacy considerations. To mitigate this, some system architectures use ASR-based clues from the full context of the audio as compared to the wake-word detector which only focuses on hypothesized wake-word segment of the audio. ASR lattice-based models have successfully been explored in this direction~\cite{Ladhak2016,jeon2019latrnn,haung2019study,dighe2020latticebased,dighe2020knowledge,agarwal2020complementary,Mallidi2018}, showing that confusion in the ASR lattices provides a strong signal of falsely accepting unintended speech. In this work, we propose a novel approach to the user's intent classification that detects if a given speech utterance accepted by the wake-word detector is actually intended towards the VA or not. Unlike traditional intent classification approaches which model acoustic and textual space at phoneme and word level respectively~\cite{sigtia2018vt}-\cite{dighe2020latticebased}, our audio-to-intent classification model learns robust acoustic and textual information at \textit{subword token-level}, and provides improved classification accuracy when compared to the baseline approaches.

\begin{figure}[t]
\centering
\includegraphics[width=\linewidth]{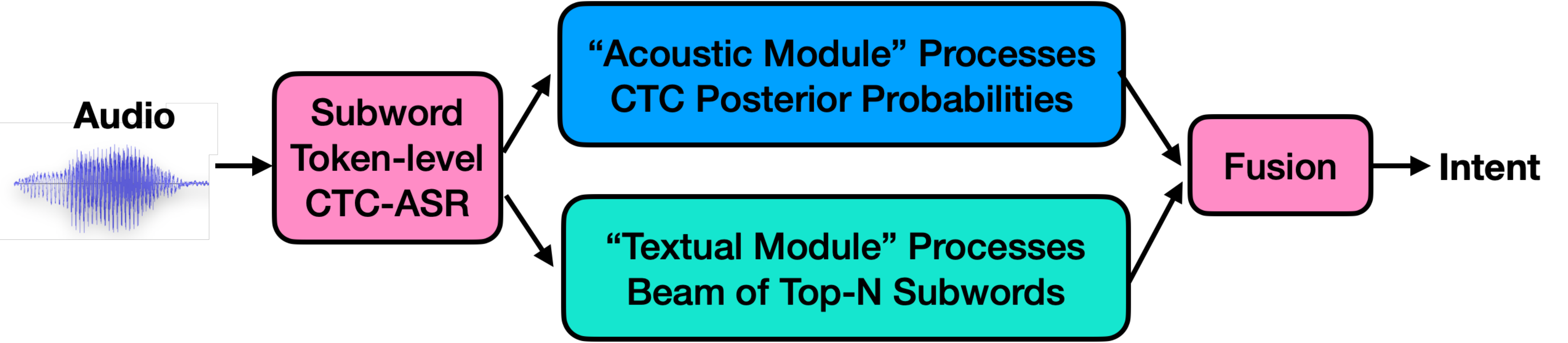}
\caption{Skeleton of our audio-to-intent approach}
\label{fig:skeleton}
\end{figure}

In our approach, shown in Figure \ref{fig:skeleton}, an end-to-end ASR model directly predicts the frame-level subword-token probabilities from the audio. An {\it acoustic} module processes these subword-token posteriors as a \textit{sum-of-posteriors} vector which is obtained by the \textit{logsumexp} operation. While we discard the frame-level granularity of token probabilities, the sum-of-posteriors vector still captures the acoustic content of the utterance as well as the uncertainty in ASR when predicting the correct tokens. Such sum-of-posteriors vectors were recently shown to be informative for training an ASR model without knowing the order of the words \cite{pratap2022wordorder}.  The goal of the acoustic module is to process acoustic information in the audio, without explicitly modeling the semantics of the user's speech. For example, for a given query \textit{``what is deep learning''}, the sum-of-posteriors vectors would contain high probabilities for subword tokens \textit{``\_ what'', ``\_ is'', ``\_ deep'', ``\_ learn'', ``ing''} as well non-zero probabilities for other subword tokens that are (typically) confused (e.g. \textit{``\_ yearn''} subword from the word \textit{``yearning''}). To account for speech semantics, our architecture is comprised of a dedicated {\it textual} module that processes a beam of top-N subword-tokens predicted at each output frame, and models the (contextual) semantic information in the resulting subword-token sequences. Specifically, each subword-token is first represented by a pretrained continuous bag-of-words token-level embedding \cite{mikolov2013cbow}, which are further augmented with mean positional-encodings that captures the sequential and multiple-hypotheses information from ASR. Finally, we fuse the acoustic and textual representations learned from the two submodules to train our audio-to-intent (A2I) classifier. 

The main contributions of this work can be summarized as follows. First, we propose a novel approach that combines multiple sources of information (acoustics and text) under a common subword-level representation learning. Second, we use sum-of-posterior vectors that capture a multinomial probability distribution over the subword token space. The entropy of the sum-of-posteriors vectors encodes the uncertainty of ASR in decoding the underlying audio. As we show in our experiments, this acoustic representation contains useful discriminative information for the task of intent classification. Third, the textual module encodes information from multiple ASR hypotheses when it processes a beam of Top-N tokens per frame, thus producing a much richer and more robust representation of the audio than processing only the top ASR hypothesis. We demonstrate the effectiveness of this approach in the challenging task of audio-to-intent classification. To the best of our knowledge this is the first approach that combines acoustic and text information at subword-level for the task of intent classification.




The rest of the paper is organized as follows. Section \ref{sec:features} discusses our novel intent classification approach. Section \ref{sec:experiments} provides details of the datasets, metrics, experimental results, and subsequent analysis. 
Finally, in Section \ref{sec:conclusions} we draw conclusions and outline future work.

\begin{figure}[t]
\begin{center}
\includegraphics[width=0.9\linewidth]{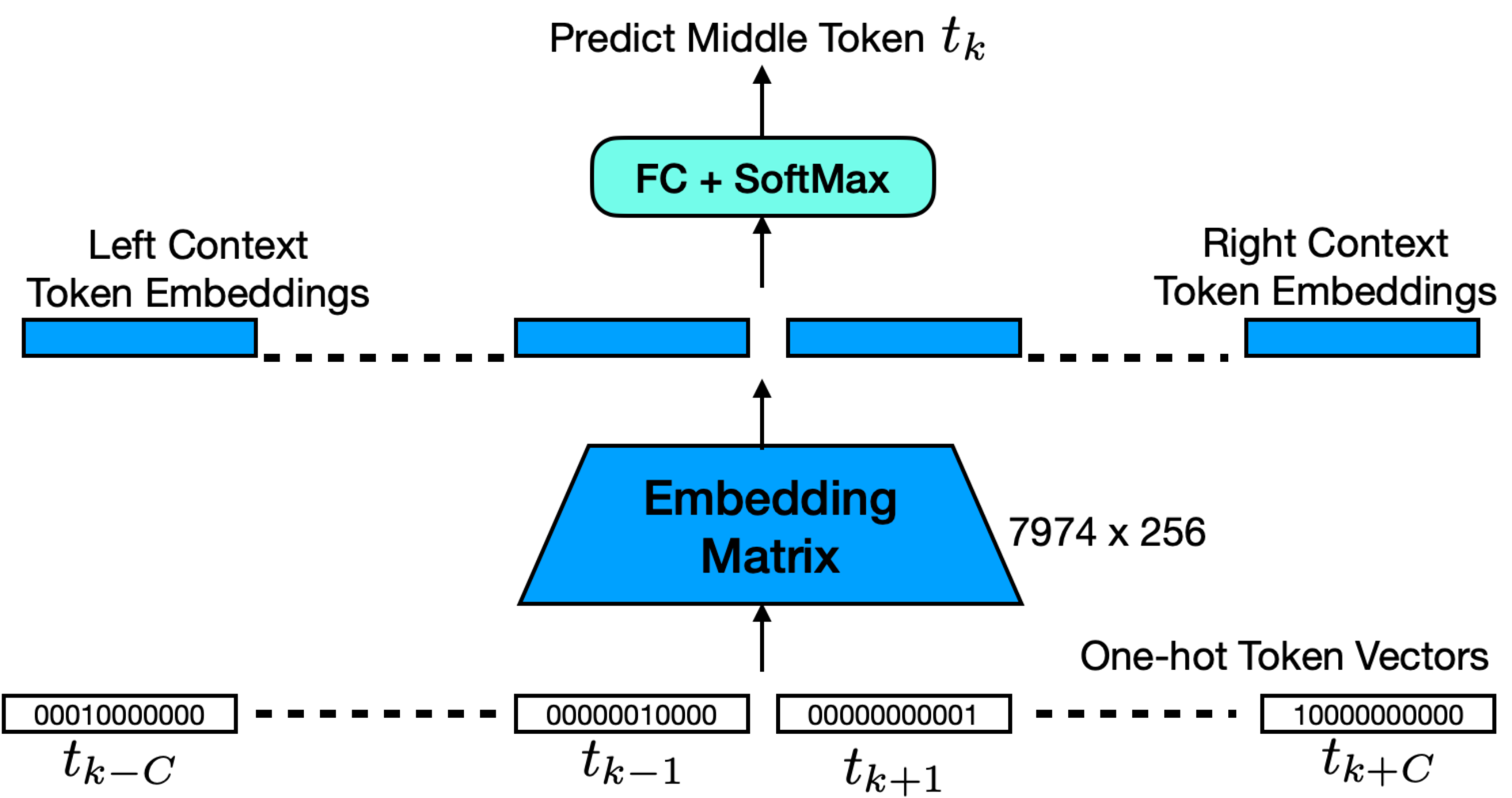}
\end{center}
\caption{Continuous bag-of-words model for learning subword token-level embeddings}
\label{fig:cbow}
\end{figure}
\vspace{-2mm}
\section{Our Approach}
\vspace{-2mm}
\label{sec:features}

\subsection{Subword Token Space for Modeling Speech and Text}
\vspace{-2mm}
To classify the user's intent accurately, analyzing acoustic differences between intended/unintended audio alone is often not sufficient; it also requires understanding the semantic meaning of the spoken sentence. Typical speech processing units like phonemes provide effective means of capturing target acoustic differences and have a fixed and relatively small vocabulary size ($\sim$40-50). Yet, these representations do not preserve the semantic meaning associated with the sequence of words/sentence. On the other hand, while using the whole words vocabulary provides richer semantic context, the complete set of words in an open vocabulary application is often too large (e.g., $\sim$600k in English language). This poses serious challenges for speech processing on time and memory constrained applications. Therefore, to mitigate the limitations of these two approaches, we propose a solution that uses a subword token space for modeling speech as well as text in our work. We deploy the SentencePiece tokenizer~\cite{kudo-richardson-2018-sentencepiece} to this end. Compared to monophones, the benefit of using the subword tokens comes from the fact that they have loose semantic meaning that can be modeled by learning token embeddings. Also, subwords tokens have a limited vocabulary size (7,974 in our case) which makes them computationally much more effective compared to the full word vocabulary which is $\sim$75 times larger.

\begin{figure}[h!t]
\begin{center}
\includegraphics[width=\linewidth]{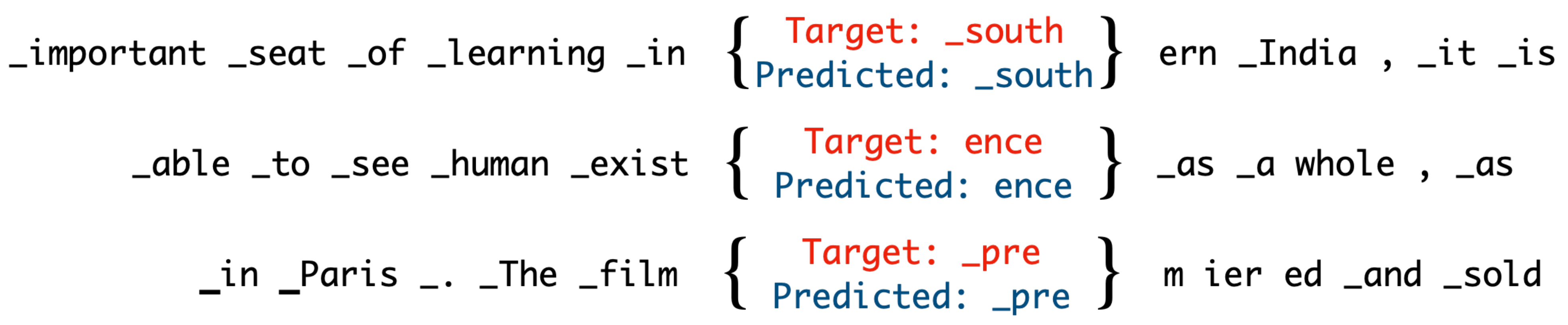}
\end{center}
\caption{Example predictions from the token-level CBOW model.}
\label{fig:cbow_example}
\end{figure}

\begin{figure*}[h!t]
\centering
\includegraphics[width=0.85\linewidth]{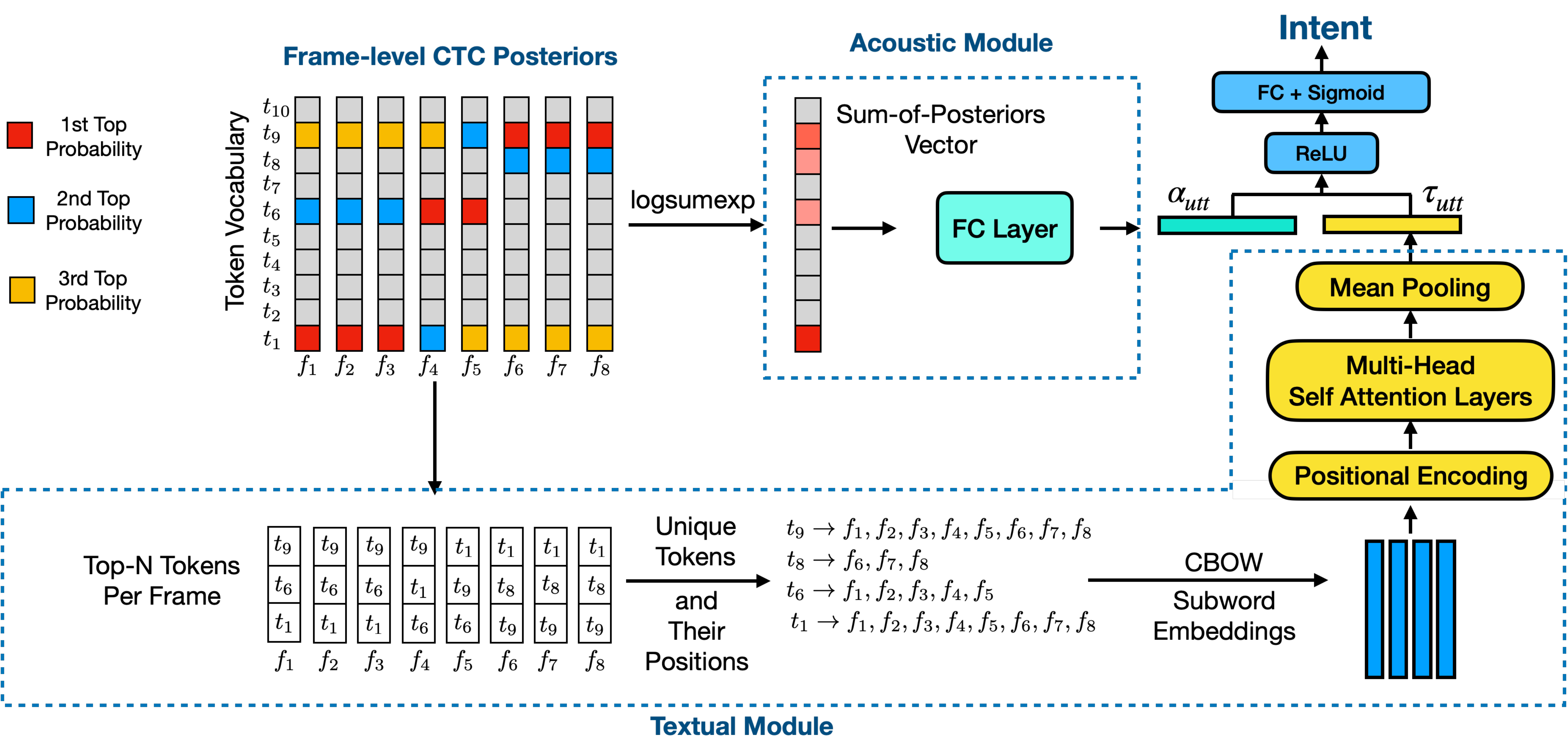}
\caption{Architecture of the proposed Audio-to-Intent approach.}
\label{fig:approach}
\end{figure*}

\subsection{Subword Token Embeddings using Continuous BOWs }
\label{sec:sentpiece}
For learning subword token-level embeddings, we adopt the continuous bag-of-words (CBOW) approach~\cite{mikolov2013cbow}. The CBOW model is trained on the publicly available Wikipedia English dataset~\cite{wikidump}. We tokenize the Wikipedia data into subword token sequences and create $\sim$181M token-level bag-of-words for training the CBOW model. We employ the masking approach where each bag-of-words contains five past and five future tokens to predict the middle token. This is depicted in Figure \ref{fig:cbow}, where the embedding matrix projects each subword token to a 256-dim embedding. The bag-of-subwords is then used to predict the middle token. The training consists of 20 epochs, and uses the cross-entropy loss for predicting the target (middle) subword. We use the learned embedding matrix as a lookup table in experiments later. Figure \ref{fig:cbow_example} shows some examples of token level predictions made using the CBOW model. As shown, five subword tokens may correspond to a context of 1 to 5 full words.

\subsection{Subword Token Posteriors from End-to-end ASR}
\label{sec:e2e_asr}
The employed E2E ASR model is based on the Conformer encoder~\cite{conformer} architecture and it is trained using CTC loss~\cite{graves2006connectionist,watanabe2018espnet} to predict frame-level posterior probabilities of the SentencePiece subword tokens, as described in Section \ref{sec:sentpiece}. Conformer encoders have recently been proposed for speech recognition tasks~\cite{yao2021wenet, zhang2022wenet,wu2021u2++}. The encoder consists of 12 Conformer layers, which have alternate transformer and convolutional layers for capturing the global and local task-specific context in audio. In our model architecture, we use the same hyper-parameter settings as mentioned in \cite{wu2021u2++}, resulting in $\sim$90M parameters in total. We train the model for 100 epochs on $\sim$18k hours of speech data. 
Once trained, we freeze this model and use it to generate subword token-level posteriors for any target utterance.
Therefore, the employed E2E ASR model acts as a feature extractor for the intent classification approach.
\vspace{-2mm}
\subsection{Joint Acoustic-Textual Modeling for Intent Classification}
The overall architecture of our intent classification approach is shown in Figure~\ref{fig:approach}. We start by representing each audio frame ($f^{th}$) with a logarithmic scale frame-level CTC posteriors ($l_f$), obtained from the E2E ASR model. The CTC posteriors are often sparse and have high probabilities only in a few token dimensions. Instead of processing the large sparse matrix of size $T\times F$, where $T=7974$ is the subword-vocabulary size and $F$ is the number of audio frames, we compress the CTC posteriors into a sum-of-posteriors (SoP) vector using logsumexp operation and normalization by the number of frames $F$ as follows:
\begin{equation}
     SoP_{utt} = logsumexp(l_1, l_2, \ldots, l_F) - log(F)
\end{equation}
The $T$-dim ${SoP}_{utt}$ vector is then transformed to a dense acoustic embedding $\alpha_{utt}$ by processing it through a fully-connected layer.\footnote{While we denote processing of the $SoP_{utt}$ vector as "acoustic" module in this work, we recognize that frame-level posteriors obtained for subword token space may benefit from some implicit language modeling information learned from the global context seen by the Conformer layers.}.

For text processing, we are primarily interested in encoding information from the top few hypotheses of the ASR output, which is in contrast to traditional text-based models that focus only on the top hypothesis, which in turn may not be optimal as it is prone to ASR errors. An obvious choice for obtaining multiple competing ASR hypotheses is Beam-Search decoding~\cite{cmu1977speech}, but in the absence of any token transition probabilities and auto-regressive decoding mechanism, we choose instead a relatively \textit{simple} and fast ``Top-N tokens per frame'' strategy. Specifically, we first take the Top-N subword tokens from each frame-level posterior, where N is a tunable parameter of the model. Then, to preserve sequential information for each unique token in the beam of Top-N subword tokens, we note down the frame-wise positions where the target token was observed. The \textit{``textual module''} block in Figure \ref{fig:approach} depicts this process. If token $t_n$ was observed in frames $f_i$, $f_j$, and $f_k$, then we construct the contextual token embedding $E_{t_n}$ as follows:

\begin{equation}
     \begin{split}
     E_{t_n} = CBOW(t_n) + mean(PE(f_i), PE(f_j), PE(f_k)),
     \end{split}
     \label{eq:emb}
\end{equation}

where $CBOW(t_n)$ represents pretrained CBOW embedding for $t_n$ and $PE(f_i)$ represents the positional encoding capturing the absolute position of the tokens in the sequence using sine and cosine functions of different frequencies~\cite{vaswani2017attention}. We take the mean\footnote{Adding two sinusodial waves of different frequencies does not result in a sinusodial wave with a modified frequency. Particularly,
{\scriptsize $\sin(\theta) + \sin(\phi) = 2\sin(\frac{\theta+\phi}{2})\cos(\frac{\theta-\phi}{2})$}
which implies that {\scriptsize $mean(PE(f_i), PE(f_j))\neq PE(mean(f_i, f_j))$}. We hypothesize that the mean positional encodings in \eqref{eq:emb} represent unique features which encode multiple positions where a given token appears. The new positional encodings do not have any direct correspondence with a linear sequence of associated “indices” as proposed in~\cite{vaswani2017attention} and simply follow a \textit{bag-of-positions} interpretation as discussed in \cite{bag_of_positions}. Eq.\eqref{eq:emb} also ensures that the numerical range of $E_{t_n}$ is consistent irrespective of how many frames a token appears in.} of all the positional encodings where $t_n$ was observed and add it to the static CBOW embedding to encode the positional information about the target token in its final representation $E_{t_n}$. If there are $K$ unique tokens in an utterance, we now have a matrix of $K\times D$ size, where $D$ is the CBOW embedding size. We process this input text representation using multiple layers of multi-headed self-attention (SA) followed by the mean-summarization layer that aggregates information across the $K$ tokens. The aggregated text representation is stored as the embedding $\tau_{utt}$. Once we have derived the acoustic and text representations, as described above, we fuse them into a single representation by concatenating the acoustic and text embeddings. These were then used as input to our audio-to-intent classifier, which is comprised of a simple non-linear (ReLU) and a fully connected layer trained to make binary decisions for the target task, as depicted in Figure~\ref{fig:approach}.

\vspace{-2mm}
\section{Experiments}
\label{sec:experiments}
\vspace{-2mm}
\subsection{Datasets, Evaluation Metrics, and Models}
\label{sec:data}
Table \ref{tab:dataset} summarizes the two-class intent classification dataset used in our experiments. This dataset is disjoint from the datasets which are used to train the CBOW token embedding model and the E2E ASR model. As the intent classification dataset is gated by an initial wake-word detection mechanism, occurrence of unintended audio is uncommon, which can also be seen from the class imbalance in our datasets. The acoustic and textual modules in Figure \ref{fig:approach} are trained using the \textit{train} set, and model convergence is tracked on the \textit{validation} partition. All variations of the intent classifier are evaluated on \textit{eval} dataset in Section \ref{sec:exp_results}, where we compare the equal error rate (EER) to capture the overall accuracy of our models. In practice, we expect our models to have minimal false alarms and maximum true positives for a good user-experience. Therefore, we also compare false alarm rate (FAR) of our models at a fixed operating point of high true positive rate (TPR) equal to 0.99. 

\begin{table}[b]
\small
\centering
\begin{tabular}{lccc}
\hline 
Class & train & validation & eval \\ 
\hline 
Intended & 90,634 & 9,809 & 27,338 \\
Unintended & 21,421 & 2,261 & 1,609 \\
\hline 
\end{tabular} 
\caption{Dataset for intent classification task.}
\label{tab:dataset}
\end{table}

We experimented with different sizes for acoustic embeddings $\alpha_{utt}$ in the acoustic module and different numbers of self-attention layers in the textual module and found 512-dim acoustic embeddings and 6 SA layers in the textual module to be the optimal choice. We also perform the following ablation studies in Section \ref{sec:exp_results}: (i) in the textual module, we try different values of N - the numbers tokens per frame to construct the set of unique high probability tokens in the utterance, (ii) we train the textual module with and without position encodings for the subword tokens, and (iii) we use only acoustic or only textual module for the intent classification task and compare with the joint acoustic-textual modeling.

We compare our approach against a baseline LatticeRNN intent classification approach~\cite{Ladhak2016,jeon2019latrnn}, which uses WFST~\cite{mohri2002weighted} lattices obtained from a hybrid ASR model. For consistency, we use the same Conformer encoder from Section \ref{sec:e2e_asr} as the acoustic model in the hybrid ASR system and employ an external n-gram language model (LM) to generate full-word-level lattices. 
The baseline LatticeRNN has the same model architecture as in \cite{jeon2019latrnn} and it uses phonetic embeddings to represent words on the lattice arcs. This baseline model is also trained using datasets from Table \ref{tab:dataset}. LatticeRNN exploits the fact that lattices from intended speech have fewer competing hypotheses within them due to in-domain language and high signal-to-noise ratio, whereas noisy out-of-domain unintended speech lattices often have higher ASR uncertainty represented by multiple competing paths in the lattice. LatticeRNN has similarities with the approach presented in this paper since both of them work with an intermediate ASR representation to perform intent classification. Yet, the approach proposed here has some notable differences: (i) it avoids expensive WFST decoding with external LM by directly using raw CTC posteriors instead of word-level lattices, (ii) it uses subword-token level modeling via CBOW embeddings as compared to phonetic word-embeddings in LatticeRNN, and (iii) it uses positional encodings to capture sequential information.

\vspace{-2mm}
\subsection{Results and Analysis}
\label{sec:exp_results}
\begin{table}[t]
\centering
\begin{tabular}{c|c|c}
\hline 
\multirow{2}{*}{\shortstack{System}} & \multicolumn{2}{c}{EER(\%)}\\ 
\cline{2-3}
& Without PosEnc & With PosEnc\\
\hline 
TextualA2I (N=1) & 10.2 & 9.4\\
TextualA2I (N=3) & 4.2 & 3.1\\
TextualA2I (N=5) & 3.5 & 2.9\\
TextualA2I (N=7) & 3.2 & \textbf{2.7}\\
TextualA2I (N=9) & 4.2 & 3.2\\
\hline
\end{tabular} 
\caption{Evaluation of textual module when trained with different number of tokens selected per frame and trained with v/s without positional encodings for the subword token sequences.}
\label{tab:analysis1}
\end{table}

\begin{table}[b]
\centering
\begin{tabular}{lccc}
\hline
System & EER(\%) & FAR at TPR=0.990\\
\hline
AcousticA2I & 2.7 & 0.072\\
TextualA2I (N=7)& 2.7 & 0.078\\
FullA2I (N=1)& \textbf{2.1} & \textbf{0.067}\\
LatticeRNN & 3.5 & 0.111\\
\hline
\end{tabular}
\caption{Evaluation of various A2I systems on \textit{eval} set.} 
\label{tab:analysis2}
\end{table}

First, we perform an ablation study only on the textual module (\textit{``TextualA2I''}). Table \ref{tab:analysis1} presents the performance of this study under various configurations. We observe that using positional encodings consistently boosts our model accuracy for fixed values of N, thereby, showing the importance of adding contextual information to the CBOW token embeddings.  We also find that the EER of our models significantly improves as the textual module accesses a broader beam (N) of tokens at each frame. This confirms our hypothesis that discriminating information between intended and unintended speech can be extracted by processing multiple potential ASR hypotheses simultaneously whereas information available in the top hypothesis can often be inaccurate due to ASR errors. In Table \ref{tab:analysis1}, the \textit{TextualA2I (N=1)} system which has the highest EER uses a greedy ASR decoding where we pick only the best subword token at each frame and therefore work only with a single ASR hypothesis in the textual module. We note that there are diminishing returns at higher values of N and the model with N=7 gives the best EER of $2.7\%$.

Next, we trained the proposed audio-to-intent (A2I) classifier using only the acoustic module (\textit{``AcousticA2I''}) and it obtained an EER of 2.7\% (shown in Table \ref{tab:analysis2}) using 512-dim acoustic embeddings. Thus, we verify the hypothesis that the sum-of-posterior vectors are independently highly informative for the intent classification task. While the best AcousticA2I and the best TextualA2I models have similar performance, we emphasize that these models are trained on different feature representations derived from the same subword token-level posteriors as shown in Figure \ref{fig:approach}.

Finally, we trained a \textit{``FullA2I''} model which uses both acoustic and textual modules. Table \ref{tab:analysis2} provides a comparison of our best FullA2I model with the baseline LatticeRNN approach as well as the best AcousticA2I and TextualA2I models that we discussed before. Interestingly, the best FullA2I model that we trained required only N=1 tokens per frame in the textual module. This suggests that the intent-related information contained in the tokens beyond the greedy ASR hypothesis in the textual module can be substituted with the information available in the sum-of-posteriors vector of the acoustic module. While we end up discarding all \textit{non-top} tokens at each frame in the textual module with N=1, the utterance-level posterior probabilities of all the tokens are still available to the FullA2I model via the sum-of-posteriors vector. We observe that the FullA2I model outperforms both the best AcousticA2I and the best TextualA2I models, which demonstrates that there is complementary information in the acoustic and textual embeddings. The baseline LatticeRNN model has an EER of 3.5\% and it mitigates $\sim89\%$ of unintended utterances at the high TPR operating point of 0.990. Contrastingly, the FullA2I model has a lower EER of $2.1\%$ and it correctly mitigates $\sim93\%$ false alarms at the same TPR, thus reducing the false alarm rate by $\sim40\%$ relative. All variations of the proposed A2I approach in Table \ref{tab:analysis2} outperform the baseline LatticeRNN model. As mentioned before, the proposed approach benefits from directly processing \textit{rich} information available in the raw CTC posteriors and avoids expensive WFST decoding to obtain full-word lattices.


To gain further insights into our approach, we inspected the input feature space of the acoustic and the textual module. We generated sum-of-posteriors vectors for the \textit{validation} set and found that the average entropy of the $SoP_{utt}$ vector for intended utterances was $0.51$ and $0.60$ for intended and unintended utterances, respectively. Higher entropy for unintended utterances supports our hypothesis that ASR is more uncertain in decoding unintended speech. We also inspected the 100 most frequently predicted tokens in intended speech versus unintended speech, and found that the two sets have only $69$ common tokens and there are $31$ tokens unique to each of intended and unintended speech. We expect that these unique tokens capture the distinctions between intended and unintended speech. 



\vspace{-3mm}
\section{Conclusions}
\vspace{-3mm}
\label{sec:conclusions}
In this paper, we addressed a binary intent classification task to improve the accuracy of voice-assistants in detecting intended speech, thereby making them less intrusive and privacy preserving. We proposed a novel architecture which models acoustic and textual information at subword-token level using sum-of-posteriors vector and semantic embeddings obtained from a continuous bag-of-words model. The backbone of our architecture is an E2E ASR model which predicts subword token probabilities at the frame level. We showed the efficacy of our approach by significantly improving the intent classification performance over a baseline LatticeRNN model. We demonstrated the importance of combining evidence from multiple ASR hypotheses when textual information alone is being processed for the intent classification task. In the future, we plan to explore a streaming version of the proposed method and tackle a more challenging intent classification task in the  wakeword-free unconstrained speech scenario. We also aim to explore end-to-end joint training of the ASR and the proposed A2I approach.

\vspace{-2mm}
{\footnotesize 
\section{Acknowledgements}
\vspace{-2mm}
We thank Mingbin Xu, Sicheng Wang, Zhen Huang and Zhihong Lei for providing support on the E2E ASR model, and Russ Webb, Dogan Can, Yi Su, and Tatiana Likhomanenko for their feedback on the paper.}

\vfill\pagebreak

{\footnotesize
\bibliographystyle{IEEEbib}
\bibliography{refs}}

\begin{thebibliography}{10}

\bibitem{sigtia2018vt}
Siddharth Sigtia, Rob Haynes, Hywel Richards, Erik Marchi, and John Bridle,
\newblock ``Efficient voice trigger detection for low resource hardware,''
\newblock in {\em Proc. Interspeech 2018}, 2018, pp. 2092--2096.

\bibitem{kumatani2017direct}
Kenichi Kumatani, Sankaran Panchapagesan, Minhua Wu, Minjae Kim, Nikko Strom,
  Gautam Tiwari, and Arindam Mandai,
\newblock ``Direct modeling of raw audio with {DNN}s for wake word detection,''
\newblock in {\em 2017 IEEE Automatic Speech Recognition and Understanding
  Workshop (ASRU)}. IEEE, 2017, pp. 252--257.

\bibitem{wu2018monophone}
M.~{Wu}, S.~{Panchapagesan}, M.~{Sun}, J.~{Gu}, R.~{Thomas}, S.~N. {Prasad
  Vitaladevuni}, B.~{Hoffmeister}, and A.~{Mandal},
\newblock ``Monophone-based background modeling for two-stage on-device wake
  word detection,''
\newblock in {\em 2018 IEEE International Conference on Acoustics, Speech and
  Signal Processing (ICASSP)}, April 2018, pp. 5494--5498.

\bibitem{guo2018time}
Jinxi Guo, Kenichi Kumatani, Ming Sun, Minhua Wu, Anirudh Raju, Nikko
  Str{\"o}m, and Arindam Mandal,
\newblock ``Time-delayed bottleneck highway networks using a {DFT} feature for
  keyword spotting,''
\newblock in {\em 2018 IEEE International Conference on Acoustics, Speech and
  Signal Processing (ICASSP)}. IEEE, 2018, pp. 5489--5493.

\bibitem{norouzian2019exploring}
Atta Norouzian, Bogdan Mazoure, Dermot Connolly, and Daniel Willett,
\newblock ``Exploring attention mechanism for acoustic-based classification of
  speech utterances into system-directed and non-system-directed,''
\newblock in {\em ICASSP 2019-2019 IEEE International Conference on Acoustics,
  Speech and Signal Processing (ICASSP)}. IEEE, 2019, pp. 7310--7314.

\bibitem{Ladhak2016}
Faisal Ladhak, Ankur Gandhe, Markus Dreyer, Lambert Mathias, Ariya Rastrow, and
  Bjorn Hoffmeister,
\newblock ``Lattice {RNN}: Recurrent neural networks over lattices,''
\newblock in {\em Interspeech 2016}, 2016.

\bibitem{jeon2019latrnn}
W.~{Jeon}, L.~{Liu}, and H.~{Mason},
\newblock ``Voice trigger detection from {LVCSR} hypothesis lattices using
  bidirectional lattice recurrent neural networks,''
\newblock in {\em ICASSP 2019 - 2019 IEEE International Conference on
  Acoustics, Speech and Signal Processing (ICASSP)}, May 2019, pp. 6356--6360.

\bibitem{haung2019study}
Che-Wei Haung, Roland Maas, Sri~Harish Mallidi, and Björn Hoffmeister,
\newblock ``A study for improving device-directed speech detection toward
  frictionless human-machine interaction,''
\newblock in {\em Proc. Interspeech}, 2019.

\bibitem{dighe2020latticebased}
P.~{Dighe}, S.~{Adya}, N.~{Li}, S.~{Vishnubhotla}, D.~{Naik}, A.~{Sagar},
  Y.~{Ma}, S.~{Pulman}, and J.~{Williams},
\newblock ``{Lattice-Based Improvements for Voice Triggering Using Graph Neural
  Networks},''
\newblock in {\em Proc. ICASSP 2020}, 2020.

\bibitem{dighe2020knowledge}
Pranay Dighe, Erik Marchi, Srikanth Vishnubhotla, Sachin Kajarekar, and Devang
  Naik,
\newblock ``{Knowledge Transfer for Efficient On-device False Trigger
  Mitigation},''
\newblock in {\em Proc. ICASSP 2020}.

\bibitem{agarwal2020complementary}
Rishika Agarwal, Xiaochuan Niu, Pranay Dighe, Srikanth Vishnubhotla, Sameer
  Badaskar, and Devang Naik,
\newblock ``{Complementary Language Model and Parallel Bi-LRNN for False
  Trigger Mitigation},''
\newblock in {\em Proc. Interspeech 2020}.

\bibitem{Mallidi2018}
Sri~Harish Mallidi, Roland Maas, Spyros Matsoukas, and Bjorn Hoffmeister,
\newblock ``Device-directed utterance detection,''
\newblock in {\em Interspeech 2018}, 2018.

\bibitem{pratap2022wordorder}
Vineel Pratap, Qiantong Xu, Tatiana Likhomanenko, Gabriel Synnaeve, and Ronan
  Collobert,
\newblock ``Word order does not matter for speech recognition,''
\newblock in {\em ICASSP 2022 - 2022 IEEE International Conference on
  Acoustics, Speech and Signal Processing (ICASSP)}, 2022, pp. 7202--7206.

\bibitem{mikolov2013cbow}
Tomas Mikolov, Kai Chen, G.s Corrado, and Jeffrey Dean,
\newblock ``Efficient estimation of word representations in vector space,''
\newblock {\em Proceedings of Workshop at ICLR}, vol. 2013, 01 2013.

\bibitem{kudo-richardson-2018-sentencepiece}
Taku Kudo and John Richardson,
\newblock ``{S}entence{P}iece: A simple and language independent subword
  tokenizer and detokenizer for neural text processing,''
\newblock in {\em Proceedings of the 2018 Conference on Empirical Methods in
  Natural Language Processing: System Demonstrations}, Brussels, Belgium, Nov.
  2018, pp. 66--71, Association for Computational Linguistics.

\bibitem{wikidump}
Wikimedia Foundation,
\newblock ``Wikimedia downloads,''
\newblock Available at \url{https://dumps.wikimedia.org}.

\bibitem{conformer}
Anmol Gulati, James Qin, Chung-Cheng Chiu, Niki Parmar, Yu~Zhang, Jiahui Yu,
  Wei Han, Shibo Wang, Zhengdong Zhang, Yonghui Wu, and Ruoming Pang,
\newblock ``Conformer: Convolution-augmented transformer for speech
  recognition,'' 2020.

\bibitem{graves2006connectionist}
Alex Graves, Santiago Fern{\'a}ndez, Faustino Gomez, and J{\"u}rgen
  Schmidhuber,
\newblock ``Connectionist temporal classification: labelling unsegmented
  sequence data with recurrent neural networks,''
\newblock in {\em Proceedings of the 23rd international conference on Machine
  learning}, 2006, pp. 369--376.

\bibitem{watanabe2018espnet}
Shinji Watanabe, Takaaki Hori, Shigeki Karita, Tomoki Hayashi, Jiro Nishitoba,
  Yuya Unno, Nelson {Enrique Yalta Soplin}, Jahn Heymann, Matthew Wiesner,
  Nanxin Chen, Adithya Renduchintala, and Tsubasa Ochiai,
\newblock ``{ESPnet}: End-to-end speech processing toolkit,''
\newblock in {\em Proceedings of Interspeech}, 2018, pp. 2207--2211.

\bibitem{yao2021wenet}
Zhuoyuan Yao, Di~Wu, Xiong Wang, Binbin Zhang, Fan Yu, Chao Yang, Zhendong
  Peng, Xiaoyu Chen, Lei Xie, and Xin Lei,
\newblock ``We{N}et: Production oriented streaming and non-streaming end-to-end
  speech recognition toolkit,''
\newblock in {\em Proc. Interspeech}, Brno, Czech Republic, 2021, IEEE.

\bibitem{zhang2022wenet}
Binbin Zhang, Di~Wu, Zhendong Peng, Xingchen Song, Zhuoyuan Yao, Hang Lv, Lei
  Xie, Chao Yang, Fuping Pan, and Jianwei Niu,
\newblock ``We{N}et 2.0: More productive end-to-end speech recognition
  toolkit,''
\newblock {\em arXiv preprint arXiv:2203.15455}, 2022.

\bibitem{wu2021u2++}
Di~Wu, Binbin Zhang, Chao Yang, Zhendong Peng, Wenjing Xia, Xiaoyu Chen, and
  Xin Lei,
\newblock ``U2++: Unified two-pass bidirectional end-to-end model for speech
  recognition,''
\newblock {\em arXiv preprint arXiv:2106.05642}, 2021.

\bibitem{cmu1977speech}
CMU Computer Science~Speech Group,
\newblock ``Speech understanding systems: Summary of results of the five-year
  research effort at carnegie-mellon university,'' 1977.

\bibitem{vaswani2017attention}
Ashish Vaswani, Noam Shazeer, Niki Parmar, Jakob Uszkoreit, Llion Jones,
  Aidan~N Gomez, {\L}ukasz Kaiser, and Illia Polosukhin,
\newblock ``Attention is all you need,''
\newblock {\em Advances in neural information processing systems}, vol. 30,
  2017.

\bibitem{bag_of_positions}
Vighnesh Shiv and Chris Quirk,
\newblock ``Novel positional encodings to enable tree-based transformers,''
\newblock in {\em Advances in Neural Information Processing Systems},
  H.~Wallach, H.~Larochelle, A.~Beygelzimer, F.~d\textquotesingle
  Alch\'{e}-Buc, E.~Fox, and R.~Garnett, Eds. 2019, vol.~32, Curran Associates,
  Inc.

\bibitem{mohri2002weighted}
Mehryar Mohri, Fernando Pereira, and Michael Riley,
\newblock ``Weighted finite-state transducers in speech recognition,''
\newblock {\em Computer Speech \& Language}, vol. 16, no. 1, pp. 69--88, 2002.

\end{thebibliography}

\end{document}